\documentclass[11pt,a4paper]{article}
\usepackage{times}
\usepackage{latexsym}

\usepackage[hyphens]{url}
\usepackage[hyperref]{emnlp2018}

\aclfinalcopy 

\usepackage{amsmath}
\usepackage{amsfonts}
\usepackage{graphicx}
\usepackage{color}
\usepackage{booktabs}
\usepackage{enumerate}
\usepackage{multirow}

\usepackage{arydshln}

\usepackage[english]{babel}
\usepackage[autostyle, english=american]{csquotes}
\MakeOuterQuote{"}

\newcommand\blfootnote[1]{%
  \begingroup
  \renewcommand\thefootnote{}\footnote{#1}%
  \addtocounter{footnote}{-1}%
  \endgroup
}

\title{Adversarial Over-Sensitivity and Over-Stability Strategies \\ for Dialogue Models}

\author{Tong Niu \and Mohit Bansal \\
  UNC Chapel Hill \\
  {\tt \{tongn, mbansal\}@cs.unc.edu}   }

\date{}

\begin{document}
\maketitle

\begin{abstract}
We present two categories of model-agnostic adversarial strategies that reveal the weaknesses of several generative, task-oriented dialogue models: \textit{Should-Not-Change} strategies that evaluate over-sensitivity to small and semantics-preserving edits, as well as \textit{Should-Change} strategies that test if a model is over-stable against subtle yet semantics-changing modifications. We next perform adversarial training with each strategy, employing a max-margin approach for negative generative examples. This not only makes the target dialogue model more robust to the adversarial inputs, but also helps it perform significantly better on the original inputs. Moreover, training on all strategies combined achieves further improvements, achieving a new state-of-the-art performance on the original task (also verified via human evaluation). In addition to adversarial training, we also address the robustness task at the model-level, by feeding it subword units as both inputs and outputs, and show that the resulting model is equally competitive, requires only 1/4 of the original vocabulary size, and is robust to one of the adversarial strategies (to which the original model is vulnerable) even without adversarial training.
\end{abstract}
\section{Introduction}
\label{sect:Introduction}

Adversarial\blfootnote{We publicly release all our CoNLL code and data at: {\scriptsize\url{https://github.com/WolfNiu/AdversarialDialogue}}} evaluation aims at filling in the gap between potential train/test distribution mismatch and revealing how models will perform under real-world inputs containing natural or malicious noise.
Recently, there has been substantial work on adversarial attacks in computer vision and NLP. Unlike vision, where one can simply add in imperceptible perturbations without changing an image's meaning, carrying out such subtle changes in text is harder since text is discrete in nature~\cite{jia2017adversarial}.
Thus, some previous works have either avoided modifying original source inputs and only resorted to inserting distractive sentences~\cite{jia2017adversarial}, or have restricted themselves to introducing spelling errors~\cite{belinkov2017synthetic} and adding non-functioning tokens~\cite{shalyminov2017challenging}.
Furthermore, there has been limited adversarial work on generative NLP tasks, e.g., dialogue generation~\cite{henderson2017ethical}, which is especially important because it is a crucial component of real-world virtual assistants such as Alexa, Siri, and Google Home. It is also a challenging and worthwhile task to keep the output quality of a dialogue system stable, because a conversation usually involves multiple turns, and a small mistake in an early turn could cascade into bigger misunderstanding later on.

Motivated by this, we present a comprehensive adversarial study on dialogue models -- we not only simulate imperfect inputs in the real world, but also launch intentionally malicious attacks on the model in order to assess them on both over-sensitivity and over-stability.
Unlike most previous works that exclusively focus on Should-Not-Change adversarial strategies (i.e., non-semantics-changing perturbations to the source sequence that \emph{should not change} the response), we demonstrate that it is equally valuable to consider Should-Change strategies (i.e., semantics-changing, intentional perturbations to the source sequence that \emph{should change} the response). 

We investigate three state-of-the-art models on two task-oriented dialogue datasets. Concretely, we propose and evaluate five naturally motivated and increasingly complex Should-Not-Change and five Should-Change adversarial strategies on the VHRED (Variational Hierarchical Encoder-Decoder) model~\cite{serban2017hierarchical} and the RL (Reinforcement Learning) model~\cite{li2016deep} with the Ubuntu Dialogue Corpus~\cite{lowe2015ubuntu}, and Dynamic Knowledge Graph Network with the Collaborative Communicating Agents (CoCoA) dataset~\cite{he2017learning}.

On the Should-Not-Change side for the Ubuntu task, we introduce adversarial strategies of increasing linguistic-unit complexity -- from shallow word-level errors, to phrase-level paraphrastic changes, and finally to syntactic perturbations. We first propose two rule-based perturbations to the source dialogue context, namely Random Swap (randomly transposing neighboring tokens) and Stopword Dropout (randomly removing stopwords). Next, we propose two data-level strategies that leverage existing parallel datasets 
in order to simulate more realistic, diverse noises: namely, Data-Level Paraphrasing (replacing words with their paraphrases)
and Grammar Errors (e.g., changing a verb to the wrong tense).
Finally, we employ Generative-Level Paraphrasing, where we adopt 
a neural model to automatically generate paraphrases of the source inputs.\footnote{A real example of Generative-Paraphrasing: context "\emph{You can find xorg . conf in /etc/X11 . It 's not needed unless it is . ;-) You may need to create one yourself .}" is paraphrased as "\emph{You may find xorg . conf in /etc/X11 . It 's not necessary until it is . You may be required to create one .}"}
On the Should-Change side for the Ubuntu task, we propose the Add Negation strategy, which negates the root verb of the source input, and the Antonym strategy, which changes verbs, adjectives, or adverbs to their antonyms.
As will be shown in Section~\ref{sect:Results}, the above strategies are effective on the Ubuntu task, but not on the collaborative-style, database-dependent CoCoA task. Thus for the latter, we investigate additional Should-Change strategies including Random Inputs (changing each word in the utterance to random ones), Random Inputs with Entities (like Random Inputs but leaving mentioned entities untouched), and Normal Inputs with Confusing Entities (replacing entities in an agent's utterance with distractive ones) to analyze where the model's robustness stems from.

To evaluate these strategies,
we first show that (1) both VHRED and the RL model are vulnerable to most Should-Not-Change and all Should-Change strategies, and (2) DynoNet's robustness to Should-Change inputs shows that it does not pay any attention to natural language inputs other than the entities contained in them.
Next, observing how our adversarial strategies `successfully' fool the target models, we try to expose these models to such perturbation patterns early on during training itself, where we feed adversarial input context and ground-truth target pairs as training data.
Importantly, we realize this adversarial training via a maximum-likelihood loss for Should-Not-Change strategies, and via a max-margin loss for Should-Change strategies.
We show that this adversarial training can not only make both VHRED and RL more robust to the adversarial data, but also improve their performances when evaluated on the original test set (verified via human evaluation). 
In addition, when we train VHRED on all of the perturbed data from each adversarial strategy together, the performance on the original task improves even further, achieving the state-of-the-art result by a significant margin (also verified via human evaluation).

Finally, we attempt to resolve the robustness issue directly at the model-level (instead of adversarial-level) by feeding subword units derived from the Byte Pair Encoding (BPE) algorithm~\cite{sennrich2015neural} to the VHRED model. We show that the resulting model not only reduces the vocabulary size by around $75\%$ (thus trains much faster) and obtains results comparable to the original VHRED, but is also naturally (i.e., without requiring adversarial training) robust to the Grammar Errors adversarial strategy.
\section{Tasks and Models}
\label{sect:Tasks and Models}
For a comprehensive study on dialogue model robustness, we  investigate both semi-task-based troubleshooting dialogue (the Ubuntu task) and the new important paradigm of collaborative two-bot dialogue (the CoCoA task). The former focuses more on natural conversations, while the latter focuses more on the knowledge base. Consequently, the model trained on the latter tends to ignore the natural language context (as will be shown in Section~\ref{subsect:Adversarial Results on CoCoA}) and hence requires a different set of adversarial strategies that can directly reveal this weakness (e.g., Random Inputs with Entities). Overall, adversarial strategies on Ubuntu and CoCoA reveal very different types of weaknesses of a dialogue model.
We implement two models on the Ubuntu task and one on the CoCoA task, each achieving state-of-the-art result on its respective task. 
Note that although we employ these two strong models as our testbeds for the proposed adversarial strategies, these adversarial strategies are not specific to the two models.

\subsection{Ubuntu Dialogue}
\noindent\textbf{Dataset and Task:}
The Ubuntu Dialogue Corpus~\cite{lowe2015ubuntu} 
contains $1$ million 2-person, multi-turn dialogues 
extracted from Ubuntu chat logs, used to provide and receive technical support. 
We focus on the task of generating fluent, relevant, and goal-oriented responses.

\noindent\textbf{Evaluation Method:}
The model is evaluated on F1's for both activities (technical verbs, e.g., "download", "install") and entities (technical nouns, e.g., "root", "web"). These metrics are computed by mapping the ground-truth and model responses to their corresponding activity-entity representations using the automatic procedure described in~\newcite{serban2017multiresolution}, who found that F1 is ``particularly suited for the goal-oriented Ubuntu Dialogue Corpus'' based on manual inspection of the extracted activities and entities.
We also conducted human studies on the dialogue quality of generated responses (see Section~\ref{subsect:Ubuntu Task Training Details} for setup and Section~\ref{subsect:Adversarial Strategies on Ubuntu} for results).

\noindent\textbf{Models:}
We reproduce the state-of-the-art Latent Variable Hierarchical Recurrent Encoder-Decoder (VHRED) model~\cite{serban2016building},
and a Deep Reinforcement Learning based generative model~\cite{li2016deep}.
For the VHRED model, we apply additive attention mechanism~\cite{bahdanau2014neural} to the source sequence while keeping the remaining architecture unchanged. 
For the RL-based model, we adopt the mixed objective function~\cite{paulus2017deep} and employ a novel reward: during training, for each source sequence $S$, we sample a response $G$ on the decoder side, feed the encoder with a random source sequence $S_{\textsc{r}}$ drawn from the train set, and use $- \log P(G|S_{\textsc{r}})$ as the reward. Intuitively, if $S_{\textsc{r}}$ stands a high chance of generating $G$ (which corresponds to a large negative reward), it is very likely that $G$ is dull and generic.

\subsection{Collaborative Communicating Agents}
\noindent\textbf{Dataset and Task:}
The collaborative CoCoA\footnote{\label{footnote:cocoa-code}{\scriptsize\url{https://stanfordnlp.github.io/cocoa/}}} dialogue task involves two agents that are asymmetrically primed with a private Knowledge Base (KB), and engage in a natural language conversation to find out the unique entry shared by the two KBs.
For a bot-bot chat of the CoCoA task, a bot is allowed one of the two actions each turn: performing an UTTERANCE action, where it generates an utterance, or making a SELECT action, where it chooses an entry from the KB. Note that each bot's SELECT action is visible to the other bot, and each is allowed to make multiple SELECT actions if the previous guess is wrong.

\noindent\textbf{Evaluation Method:}
One of the major metrics is Completion Rate, the percentage of two bots successfully finishing the task.

\noindent\textbf{Models:} We focus on DynoNet, 
the best-performing model for the CoCoA task~\cite{he2017learning}. 
It consists of a dynamic knowledge graph, 
a graph embedding over the entity nodes, and a Seq2seq-based utterance generator.
\section{Adversarial Strategies}
\subsection{Adversarial Strategies on Ubuntu}
For Ubuntu, we introduce adversarial strategies of increasing linguistic-unit complexity -- from shallow word-level errors such as Random Swap and Stopword Dropout, to phrase-level paraphrastic changes, and finally to syntactic Grammar Errors.

\paragraph{Should-Not-Change Strategies\\}
\label{para:Should-Not-Change Strategies}
\noindent\textbf{(1) Random Swap:}
Swapping adjacent words occurs often in the real world, e.g., transposition of words is one of the most frequent errors in manuscripts~\cite{headlam1902transposition,marques2014errors}; it is also frequently seen in blog posts.\footnote{E.g., "\emph{he would give \textbf{to it} me}" in {\scriptsize\url{https://tinyurl.com/y92f6gz9}}} Thus, being robust to swapping adjacent words is useful for chatbots that take typed/written text as inputs (e.g., virtual customer support on a airline/bank website). Even for speech-based conversations, non-native speakers can accidentally swap words due to habits formed in their native language (e.g., SVO in English vs. SOV in Hindi, Japanese, and Korean).
Inspired by this, we also generate globally contiguous but locally "time-reversed" text, where positions of neighboring words are swapped (e.g., "\emph{I don't want you to go}" to "\emph{I don't want to you go}").

\noindent\textbf{(2) Stopword Dropout:}
Stopwords are the most frequent words in a language. The most commonly-used $25$ words in the Oxford English corpus make up one-third of all printed material in English, and these words consequently carry less information than other words do in a sentence.\footnote{One could also use closed-class words (prepositions, determiners, coordinators, and pronouns), but we opt for stopwords because a majority of stopwords are indeed closed-class words, and secondly, closed-class words usually require a very accurate POS-tagger, which is not available for low-resource or noisy domains and languages (e.g., Ubuntu).}
Inspired by this observation, we propose randomly dropping stopwords from the inputs (e.g., "\emph{Ben ate the carrot}" to "\emph{Ben ate carrot}").

\noindent\textbf{(3) Data-level Paraphrasing:}
We repurpose PPDB $2.0$~\cite{pavlick2015ppdb} and replace words and phrases in the original inputs with their paraphrases (e.g., "\emph{She bought a bike}" to "\emph{She purchased a bicycle}").

\noindent\textbf{(4) Generative-level Paraphrasing:} 
Although Data-level Paraphrasing provides us with semantic-preserving inputs most of the time,
it still suffers from the fact that the validity of a paraphrase depends on the context, 
especially for words with multiple meanings. In addition, simply replacing word-by-word does not lead to new compositional sentence-level paraphrases, e.g., "\emph{How old are you}" to "\emph{What's your age}".
We thus also experiment with generative-level paraphrasing, 
where we employ the Pointer-Generator Networks~\cite{see2017get}, 
and train it on the recently published paraphrase dataset ParaNMT-5M~\cite{wieting2017pushing} which contains $5$ millions paraphrase pairs.

\noindent\textbf{(5) Grammar Errors:}
We repurpose the AESW dataset~\cite{daudaravicius2015automated}, text extracted from $9,919$ published journal articles
with data before/after language editing. This dataset was used for training models that identify and correct grammar errors.
Based on the corrections
in the edits, 
we build a look-up table to replace each correct word/phrase with a wrong one
(e.g., "\emph{He doesn't like cakes}" to "\emph{He don't like cake}").

\paragraph{Should-Change Strategies\\}
\noindent\textbf{(1) Add Negation:}
Suppose we add negation to the source sequence of some task-oriented model --- from "\emph{I want some coffee}" to "\emph{I don't want some coffee}". 
A proper response to the first utterance could be "\emph{Sure, I will bring you some coffee}", but for the second one, the model should do anything but bring some coffee. We thus assume that if we add negation to the root verb of each source sequence and the response is unchanged, 
the model must be ignoring important linguistic cues like negation. Hence this qualifies as a Should-Change strategy, i.e., if the model is robust, it should change the response.

\noindent\textbf{(2) Antonym:}
We change words in utterances to their antonyms to apply more subtle meaning changes 
(e.g., "\emph{You need to install Ubuntu}" to "\emph{You need to uninstall Ubuntu}").\footnote{Note that Should-Change strategies may lead to contexts that do not correspond to any legitimate task completion action, but the purpose of such a strategy is to make sure that the model at least should not respond the same way as it responded to the original context, i.e., even for the no-action state, the model should respond with something different like "\emph{Sorry, I cannot help with that.}" Our semantic similarity results in Table~\ref{tab:ubuntu-similarity-result-vhred} capture this intuition directly.}

\subsection{Adversarial Strategies on CoCoA}
We applied all the above successful strategies used for the Ubuntu task to the UTTERANCE actions in a bot-bot-chat setting for the CoCoA task, but found that none of them was effective on DynoNet. This is surprising considering that the model's language generation module is a traditional Seq2seq model. This observation motivated us to perform the following analysis.
The high performance of bot-bot chat may have stemmed from two sources: information revealed in an utterance, 
or entries directly disclosed by a SELECT action.

To investigate which part the model relies on more, we experiment with different Should-Change strategies which
introduce obvious perturbations that have minimal word or semantic meaning overlap with the original source inputs:

\noindent\textbf{(1) Random Inputs:} 
Turn both bots' utterances into random inputs. This aims at investigating how much the model depends on the SELECT action.

\noindent\textbf{(2) Random Inputs with Kept Entities:} 
Replace each bot's utterance with random inputs, but keep the contained entities untouched. This further investigates how much entities alone contribute to the final performance.

\noindent\textbf{(3) Confusing Entity:} 
Replace entities mentioned in bot A's utterances with entities that are present in bot B's KB but not in their shared entry (and vice versa). This aims at coaxing bot B into believing that the mentioned entities come from their shared entry.
By intentionally making the utterances misleading, we expect DynoNet's performance to be lower -- hence this qualifies as a Should-Change strategy.

\section{Adversarial Training}
\label{sect:Adversarial Training}
To make a model robust to an adversarial strategy, a natural approach is exposing it to the same pattern of perturbation during training (i.e., \textit{adversarial training}). This is achieved by feeding adversarial inputs as training data.
For each strategy, we report results under three train/test combinations: 
(1) trained with normal inputs, tested on adversarial inputs (\textit{N-train + A-test}), which evaluates whether the adversarial strategy is effective at fooling the model and exposing its robustness issues; 
(2) trained with adversarial inputs, tested on adversarial inputs (\textit{A-train + A-test}), which next evaluates whether adversarial training made the model more robust to that adversarial attack; and 
(3) trained with adversarial inputs, tested on normal inputs (\textit{A-train + N-test}), which finally evaluates whether the adversarial training also makes the model perform equally or better on the original normal inputs.
Note that (3) is important, because one should not make the model more robust to a strategy at the cost of lower performance on the original data; also when (3) improves the performance on the original inputs, it means adversarial training successfully teaches the model to recognize and be robust to a certain type of noise, so that the model performs better when encountering similar patterns during inference.
Also note that we use perturbed train set for adversarial training, and perturbed test set for adversarial testing. There is thus no overlap between the two sets.

\subsection{Adversarial Training for Should-Not-Change Strategies}
For each Should-Not-Change strategy, we take an already trained model from a certain checkpoint,\footnote{We do not train from scratch because each model (for each strategy) takes several days to converge.
}
and train it on the adversarial inputs with maximum likelihood loss for $K$ epochs~\cite{shalyminov2017challenging,belinkov2017synthetic,jia2017adversarial,iyyer2018adversarial}. By feeding "adversarial source sequence + ground-truth response pairs" as regular positive data, we teach the model that these pairs are also valid examples despite the added perturbations.

\subsection{Adversarial Training for Should-Change Strategies}
\label{subsect:Adversarial Training for Should-Change Strategies}
For Should-Change strategies, we want the F1's to be lower with adversarial inputs after adversarial training, since this shows that the model becomes sensitive to subtle yet semantic-changing perturbations. 
This cannot be achieved by naively training on the perturbed inputs with maximum likelihood loss, because 
the "perturbed source sequence + ground-truth response pairs" for Should-Change strategies are negative examples which we need to train the model to avoid from generating.
Inspired by~\newcite{mao2016generation} and~\newcite{yu2017joint}, 
we instead use a linear combination of maximum likelihood loss and max-margin loss:
\begin{align*}
\vspace{-5pt}
	L \! &= \! L_{\textsc{ml}} \! + \! \alpha L_{\textsc{mm}} \\
    L_{\textsc{ml}} \! &= \! \sum_i \! \log \! P(t_i|s_i) \\
    L_{\textsc{mm}} \! &= \! \sum_i \! \max{(0, M \! + \! \log \! P(t_i|a_i) \! - \! \log \! P(t_i|s_i))}
\vspace{-10pt}
\end{align*}
where $L_{\textsc{ml}}$ is the maximum likelihood loss, $L_{\textsc{mm}}$ is the max-margin loss, $\alpha$ is the weight of the max-margin loss (set to $1.0$ following~\newcite{yu2017joint}), $M$ is the margin (tuned be to $0.1$), and $t_i$, $s_i$ and $a_i$ are the target sequence, normal input, and adversarial input, respectively.\footnote{Please refer to supp. about greedy sampling based max-margin setup and CoCoA discussion for adversarial training.}
\section{Experimental Setup}
\label{sect:Experimental Setup}
In addition to datasets, tasks, models and evaluation methods introduced in Section~\ref{sect:Tasks and Models}, we present training details in this section (see Appendix for a comprehensive version).

\begin{table}[t]
\small
\centering
\begin{tabular}{lcc}
\hline
Model & Activity F1 & Entity F1 \\
\midrule
LSTM & 1.18 & 0.87 \\
HRED & 4.34 & 2.22 \\
VHRED & 4.63 & 2.53 \\
VHRED (w/ attn.) & \textbf{5.94} & 3.52 \\
Reranking-RL & 5.67 & \textbf{3.73} \\ \hline
\end{tabular}
\vspace{-4pt}
\caption{F1 results of previous works as compared to our models. LSTM, HRED and VHRED are results reported in~\newcite{serban2017multiresolution}. VHRED (w/ attn.) and Reranking-RL are our results. Top results are bolded.} 
\label{tab:ubuntu-F1-result-previous-works}
\vspace{-8pt}
\end{table}

\begin{table*}[t]
\small
\centering
\begin{tabular}{lcccc}
\toprule
Strategy Name & N-train + A-test & A-train + A-test & A-train + N-test & N-train + N-test \\
\midrule
Normal Input & - & - & - & 5.94, 3.52 \\
\midrule
\midrule
Random Swap & 6.10*, 3.42\phantom{*} & 6.47*, 3.64* & 6.42*, 3.74* & - \\
Stopword Dropout & 5.49*, 3.44\phantom{*} & 6.23*, 3.82* & 6.29*, 3.71* & - \\
Data-Level Para. & 5.38*, 3.18* & 6.39*, 3.83* & 6.32*, 3.87* & - \\
Generative-Level Para. & 4.25*, 2.48* & 5.89\phantom{*},  3.60\phantom{*} & 6.11*, 3.66* & - \\
Grammar Errors & 5.60*, 3.09* & 5.93\phantom{*},  3.67* & 6.05\phantom{*},  3.69* & - \\
\hdashline
All Should-Not-Change & - & - & 6.74*, 3.97* & - \\
\midrule
\midrule
Add Negation & 6.06\phantom{*},  3.42\phantom{*} & 5.01*, 3.12* & 6.07\phantom{*},  3.46\phantom{*} & - \\
Antonym & 5.85\phantom{*},  3.56\phantom{*} & 5.43*, 3.43\phantom{*} & 5.98\phantom{*},  3.56\phantom{*} & - \\
\bottomrule
\end{tabular}
\vspace{-5pt}
\caption{Activity and Entity F1 results of adversarial strategies on the \textbf{VHRED} model. Numbers marked with * are stat. significantly higher/lower than their counterparts obtained with Normal Input (upper-right corner of table).}
\label{tab:ubuntu-F1-result-vhred}
\vspace{-8pt}
\end{table*}

\noindent\textbf{Models on Ubuntu:}
\label{subsect:Ubuntu Task Training Details}
We implemented VHRED and Reranking-RL in TensorFlow~\cite{abadi2016tensorflow}
and employed greedy search for inference.
As shown in Table~\ref{tab:ubuntu-F1-result-previous-works}, for both models we obtained Activity and Entity F1's higher than the VHRED results reported in~\newcite{serban2017multiresolution}.
Hence, each of these two implementations serves as a solid baseline for adversarial testing and training.

\noindent\textbf{Should-Not-Change Strategies on Ubuntu:}
\label{para:Should-Not-Change Strategies Setup}
For Random Swap, we allow up to $1$ swap of neighboring words per $4$ words in each utterance.
For Stopword Dropout,
we allow up to $8$ words to be dropped in each turn.
For Data-level Paraphrasing, we use the small version of PPDB 2.0. 
For Generative-level Paraphrasing, we use the publicly available Pointer-Generator Networks code
(See Appendix for some random samples of the generated paraphrases).\footnote{{\scriptsize\url{https://github.com/becxer/pointer-generator}}}
For Grammar Errors, in addition to those extracted from the AESW dataset, we also add a heuristic where an inflected verb is replaced with its respective infinitive form,
and a plural noun with its singular form. 
Note that for all strategies we only keep an adversarial token if it is within the original vocabulary set.

\noindent\textbf{Should-Change Strategies on Ubuntu:}
For Add Negation, we negate the first verb in each utterance.
For Antonym, we modify the first verb, adjective or adverb that has an antonym.

\noindent\textbf{Human Evaluation:}
We also conducted human studies on MTurk to evaluate adversarial training (pairwise comparison for dialogue quality) and generative paraphrasing (five-point Likert scale). The utterances were randomly shuffled to anonymize model identity, and we used MTurk with US-located human evaluators with approval rate $>98\%$, and at least $10,000$ approved HITs. Results are presented in Section~\ref{para:Human Evaluation}.
Note that the human studies and automatic evaluation are complementary to each other: while MTurk annotators are good at judging how natural and coherent a response is, they are usually not experts in the Ubuntu operating system's technical details. On the other hand, automatic evaluation focuses more on the technical side (i.e., whether key activities or entities are present in the response).

\noindent\textbf{Model on CoCoA:}
We adopted the publicly available code from~\newcite{he2017learning},\footnote{{\scriptsize\url{https://tinyurl.com/ydheoa8l}}} and used their already trained DynoNet model.
\vspace{-4pt}
\section{Results}
\label{sect:Results}

\vspace{-4pt}
\subsection{Adversarial Results on Ubuntu}
\label{subsect:Adversarial Strategies on Ubuntu}

\paragraph{Result Interpretation}
For Table~\ref{tab:ubuntu-F1-result-vhred} and~\ref{tab:ubuntu-F1-result-RL} with Should-Not-Change strategies, lower is better in the first column (since a successful adversarial testing strategy will be effective at fooling the model), while higher is better in the second column (since successful adversarial training should bring the performance back up).
However, for Should-Change strategies, the reverse holds.\footnote{Higher is better in the first column, because this shows that the model is not paying attention to important semantic changes in the source inputs (and is maintaining its original performance); while lower is better in the second column, since we want the model to be more sensitive to such changes after adversarial training.}  Lastly, in the third column, higher is better since we want the adversarially trained model to perform better on the original source inputs.

\begin{table*}[t]
\small
\centering
\begin{tabular}{lcccc}
\toprule
Strategy Name & N-train + A-test & A-train + A-test & A-train + N-test & N-train + N-test \\
\midrule
Normal Input & - & - & - & 5.67, 3.73 \\
\midrule
\midrule
Random Swap & 5.49*, 3.56* & 6.20*, 4.28* & 6.36*, 4.39* & - \\
Stopword Dropout & 5.51*, 4.09* & - & - & - \\
Data-Level Para. & 5.28*, 3.07* & 5.53*, 3.69\phantom{*} & 5.79*, 3.87* & - \\
Generative-Level Para. & 4.47*, 2.63* & 5.30*, 3.35* & 5.86*, 3.90* & - \\
Grammar Errors & 5.33*, 3.25* & 5.55*, 3.92* & 5.93*, 4.04* & - \\
\midrule
\midrule
Add Negation & 5.61\phantom{*}, 3.79\phantom{*} & 4.92*, 2.78* & 6.10*, 3.93* & - \\
Antonym & 5.68\phantom{*}, 3.70\phantom{*} & 5.30*, 2.95* & 5.80*, 3.71\phantom{*} & - \\
\bottomrule
\end{tabular}
\vspace{-5pt}
\caption{Activity and Entity F1 results of adversarial strategies on the \textbf{Reranking-RL} model. Numbers marked with * are stat. significantly higher/lower than their counterparts obtained with Normal Input (upper-right corner).
}
\label{tab:ubuntu-F1-result-RL}
\vspace{-7pt}
\end{table*}

\vspace{-4pt}
\paragraph{Results on Should-Not-Change Strategies}
Table~\ref{tab:ubuntu-F1-result-vhred} and~\ref{tab:ubuntu-F1-result-RL}
present the adversarial results on F1 scores of all our strategies for VHRED and Reranking-RL, respectively.
Table~\ref{tab:ubuntu-F1-result-vhred} shows that VHRED is robust to none of the Should-Not-Change strategies other than Random Swap, while Table~\ref{tab:ubuntu-F1-result-RL} shows that Reranking-RL is robust to none of the Should-Not-Change strategies other than Stopword Dropout. For each effective strategy, at least one of the F1's decreases statistically significantly\footnote{We obtained stat. significance via the bootstrap test~\cite{noreen1989computer,efron1994introduction} with 100K samples, and consider $p < 0.05$ as stat. significant.}
as compared to the same model fed with normal inputs.
Next, all adversarial trainings on Should-Not-Change strategies not only make the model more robust to adversarial inputs (each \textit{A-train + A-test} F1 is stat. significantly higher than that of \textit{N-train + A-test}) , but also make them perform better on normal inputs (each \textit{A-train + N-test} F1 is stat. significantly higher than that of \textit{N-train + N-test}, except for Grammar Errors's Activity F1).
Motivated by the success in adversarial training on each strategy alone, we also experimented with training on all Should-Not-Change strategies combined, and obtained F1's stat. significantly higher than any single strategy (the \textit{All Should-Not-Change} row in Table~\ref{tab:ubuntu-F1-result-vhred}), except that \textit{All-Should-Not-Change}'s Entity F1 is stat. equal to that of Data-Level Paraphrasing, showing that these strategies are able to compensate for each other to further improve performance.
An interesting strategy to note is Random Swap: although it itself is not effective as an adversarial strategy for VHRED, training on it does make the model perform better on normal inputs.

\begin{table}[t]
\small
\centering
\begin{tabular}{lcccc}
\toprule[0.5pt]
\multirow{2}{*}{ Strategy Name } & \multicolumn{2}{c}{ VHRED } & \multicolumn{2}{c}{ Reranking-RL } \\
\cmidrule[0.5pt]{2-5}
& Cont. & Resp. & Cont. & Resp. \\
\midrule
Random Swap & 1.00 & 0.71 & 1.00 & 0.86 \\
Stopword Dropout & 0.61 & 0.50 & 0.76 & 0.68 \\
Data-Level Para. & 0.96 & 0.58 & 0.96 & 0.74 \\
Gen.-Level Para. & 0.70 & 0.40 & 0.76 & 0.55 \\
Grammar Err. & 0.96 & 0.58 & 0.97 & 0.74 \\
\midrule
\midrule
Add Negation & 0.96 & 0.69 & 0.97 & 0.81 \\
Antonym & 0.98 & 0.66 & 0.98 & 0.74 \\
\bottomrule
\end{tabular}
\vspace{-4pt}
\caption{Textual similarity of adversarial strategies on the VHRED and Reranking-RL models. "Cont." stands for "Context", and "Resp." stands for "Response".
\label{tab:ubuntu-similarity-result-vhred}
}
\vspace{-10pt}
\end{table}

\paragraph{Results on Should-Change Strategies}
Table~\ref{tab:ubuntu-F1-result-vhred} and~\ref{tab:ubuntu-F1-result-RL} show that Add Negation and Antonym are both successful Should-Change strategies, because no change in \textit{N-train + A-test} F1 is stat. significant compared to that of \textit{N-train + N-test}, which shows that both models are ignoring the semantic-changing perturbations to the inputs.
From the last two rows of \textit{A-train + A-test} column in each table, we also see that adversarial training successfully brings down both F1's (stat. significantly) for each model, showing that the model becomes more sensitive to the context change.

\paragraph{Semantic Similarity}
In addition to F1, we also follow~\newcite{serban2017multiresolution} 
and employ cosine similarity between average embeddings of normal and adversarial inputs/responses (proposed by~\newcite{liu2016not}) 
to evaluate how much the inputs/responses change in semantic meaning (Table~\ref{tab:ubuntu-similarity-result-vhred}).
This metric is useful in three ways.
Firstly, by comparing the two columns of context similarity, we can get a general idea of how much change is perceived \textit{by each model}. For example, we can see that Stopword Dropout leads to more evident changes from VHRED's perspective than from Reranking-RL's. This also agrees with the F1 results in Table~\ref{tab:ubuntu-F1-result-vhred} and~\ref{tab:ubuntu-F1-result-RL}, which indicate that Reranking-RL is much more robust to this strategy than VHRED is. The high context similarity of Should-Change strategies shows that although we have added "not" or replaced antonyms in every utterance of the source inputs, from the model's point of view the context has not changed much in meaning.
Secondly, for each Should-Not-Change strategy, the cosine similarity of context is much higher than that of response, indicating that responses change more significantly in meaning than their corresponding contexts. Lastly, The high semantic similarity for Generative Paraphrasing also partly shows that the Pointer-Generator model in general produces faithful paraphrases.

\begin{table}[t]
\small
\centering
\begin{tabular}{cccc}
\toprule
Compared to Baseline & Win(\%) & Tie(\%) & Loss(\%) \\
\midrule
Random Swap & 49 & 19 & 32 \\
Stopword Dropout & 45 & 19 & 36 \\
Data-Level Para. & 37 & 22 & 41 \\
Generative-Level Para. & 41 & 26 & 33 \\
Grammar Errors & 41 & 29 & 30 \\
\hdashline
All Should-Not-Change & 49 & 22 & 28 \\ 
\midrule
\midrule
Add Negation & 34 & 25 & 41 \\
Antonym & 40 & 29 & 31 \\
\bottomrule
\end{tabular}
\vspace{-7pt}
\caption{Human evaluation results on comparison between VHRED baseline trained on normal inputs vs. VHRED trained on each Should-Not-Change strategy (incl. one with all Should-Not-Change strategies combined) and each Should-Change strategy for Ubuntu.}
\label{human-eval-vhred}
\vspace{-5pt}
\end{table}

\begin{table}[t]
\small
\centering
\begin{tabular}{c|cc}
\hline
& Pointer-Generator & ParaNMT-5M \\
\hline
Avg. Score & $3.26$ & $3.54$ \\ \hline
\end{tabular}
\vspace{-5pt}
\caption{Human evaluation scores on paraphrases generated by Pointer-Generator Networks and ground-truth pairs from ParaNMT-5M.}
\label{human-eval-para}
\vspace{-15pt}
\end{table}

\begin{table*}[t]
\small
\centering
\begin{tabular}{p{9.3cm}|p{5.7cm}}
\toprule
\multicolumn{1}{c|}{Context} & \multicolumn{1}{|c}{Response} \\
\midrule
\textbf{N:} ... you could save your ubuntu files and reinstall Windows , then install ubuntu as a dual boot option \_\_eou\_\_ \_\_eot\_\_ aight buddy , so how do i get that **unknown** space back\_\_eou\_\_ \newline \textbf{Random Swap:} ... you could your save ubuntu and files Windows reinstall , then install ubuntu as dual a option boot \_\_eou\_\_ \_\_eot\_\_ aight buddy , so do how i that get space **unknown** back \_\_eou\_\_ & \textbf{NN:} you can 	use the Live CD , you can install Ubuntu on the same partition as the Windows partition \_\_eou\_\_ \newline \textbf{NA:} I am using ubuntu . \_\_eou\_\_ \newline \textbf{AA:} you can use Windows XP on the Windows partition , and then install Ubuntu on the same drive \_\_eou\_\_ \\
\midrule
\bottomrule
\end{tabular}
\vspace{-5pt}
\caption{VHRED output example before and after adversarial training on the Random Swap strategy.}
\label{fig:generative-para-main-paper}
\vspace{-7pt}
\end{table*}
\vspace{-5pt}
\paragraph{Human Evaluation}
\label{para:Human Evaluation}
As introduced in Section~\ref{subsect:Ubuntu Task Training Details}, we performed two human studies on adversarial training and Generative Paraphrasing.
For the first study, Table~\ref{human-eval-vhred} indicates that models trained on each adversarial strategy (as well as on all Should-Not-Change strategies combined) indeed on average produced better responses, and mostly agrees with the adversarial training results in Table~\ref{tab:ubuntu-F1-result-vhred}.\footnote{Note that human evaluation does not show improvements with the Data-Level-Paraphrasing and Add-Negation strategies, though the latter does agree with F1 trends. Overall, we provide both human and F1 evaluations because they are complementary at judging naturalness/coherence vs. key Ubuntu technical activities/entities.}
For the second study, Table~\ref{human-eval-para} shows that on average the generated paraphrase has roughly the same semantic meaning with the original utterance, but may sometimes miss some information. Its quality is also close to that of the ground-truth in ParaNMT-5M dataset.

\paragraph{Output Examples of Generated Responses}
We present a selected example of generated responses before and after adversarial training on the Random Swap strategy with the VHRED model in Table~\ref{fig:generative-para-main-paper} (more examples in Appendix on all strategies with both models). 
First of all, we can see that it is hard to differentiate between the original and the perturbed context (\textit{N-context} and \textit{A-context}) if one does not look very closely.
For this reason, the model gets fooled by the adversarial strategy, i.e., after adversarial perturbation, the \textit{N-train + A-test} response (NA-Response) is worse than that of \textit{N-train + N-test} (NN-Response). However, after our adversarial training phase, \textit{A-train + A-test} (AA-Response) becomes better again.

\subsection{Adversarial Results on CoCoA}
\label{subsect:Adversarial Results on CoCoA}
Table~\ref{tab:mutual-result} shows the results of Should-Change strategies on DynoNet with the CoCoA task.
The Random Inputs strategy shows that even without communication, the two bots are able to locate their shared entry $82\%$ of the time by revealing their own KB through SELECT action.
When we keep the mentioned entities untouched but randomize all other tokens,
DynoNet actually achieves state-of-the-art Completion Rate, indicating that the two agents are paying zero attention to each other's utterances other than the entities contained in them. This is also why we did not apply Add Negation and Antonym to DynoNet --- if Random Inputs does not work, these two strategies will also make no difference to the performance (in other words Random Inputs subsumes the other two Should-Change strategies).
We can also see that even with the Normal Inputs with Confusing Entities strategy,
DynoNet is still able to finish the task $77\%$ of the time, and with only slightly more turns. This again shows that the model mainly relies on the SELECT action to guess the shared entry.

\begin{table}[t]
\small
  \centering
    \begin{tabular}{lcc}
    \toprule
    Strategy & \multicolumn{1}{l}{Completion Rate} & \multicolumn{1}{l}{Num. of Turns} \\
    \midrule
    Norm. Inputs & 0.94  & 16.06 \\
	Rand. Inputs & 0.82  & 22.87 \\
    Rand. w/ Entity & 0.95  & 17.19 \\
    Confusing Entity & 0.77  & 24.11 \\
    \bottomrule
    \end{tabular}
  \vspace{-5pt}
  \caption{Adversarial Results on DynoNet.}
  \label{tab:mutual-result}
  \vspace{-10pt}
\end{table}
\section{Byte-Pair-Encoding VHRED}
\label{subsect:Byte-Pair-Encoding VHRED}
Although we have shown that adversarial training on most strategies makes the dialogue model more robust, generating such perturbed data is not always straightforward for diverse, complex strategies. For example, our data-level and generative-level strategies all leverage datasets that are not always available to a language. We are thus motivated to also address the robustness task on the model-level, and 
explore an extension to the VHRED model that makes it robust to Grammar Errors even without adversarial training.

\noindent\textbf{Model Description:}
We performed Byte Pair Encoding (BPE)~\cite{sennrich2015neural} on the Ubuntu dataset.\footnote{
We employed code released by the authors on \url{https://github.com/rsennrich/subword-nmt}} 
This algorithm encodes rare/unknown words as sequences of subword units, which helps segmenting words with the same lemma but different inflections (e.g., "showing" to "show + ing", and "cakes" to "cake + s"), making the model more likely to be robust to grammar errors such as verb tense or plural/singular noun confusion.
We experimented BPE with $5$K merging operations, and obtained a vocabulary size of $5121$.

\noindent\textbf{Results:}
As shown in Table~\ref{tab:bpe-result}, BPE-VHRED achieved F1's ($5.99$, $3.66$), which is stat. equal to ($5.94$, $3.52$) obtained without BPE. To our best knowledge, we are the first to apply BPE to a generative dialogue task.
Moreover, 
BPE-VHRED achieved ($5.86$, $3.54$) on Grammar Errors based adversarial test set, which is stat. equal to the F1's when tested with normal data, indicating that BPE-VHRED is more robust to this adversarial strategy than VHRED is, since the latter had ($5.60$, $3.09$) when tested with perturbed data, where both F1's are stat. signif. lower than when fed with normal inputs.
Moreover, BPE-VHRED reduces the vocabulary size by $15$K, corresponding to $4.5$M fewer parameters. This makes BPE-VHRED train much faster. Note that BPE only makes the model robust to one type of noise (i.e. Grammar Errors), and hence adversarial training on other strategies is still necessary (but we hope that this encourages future work to build other advanced models that are naturally robust to diverse adversaries).
\begin{table}[t]
\small
\centering
\begin{tabular}{ccc}
\toprule
& VHRED & BPE-VHRED \\
\midrule
Normal Input & 5.94, 3.52 & 5.99, 3.66 \\
Grammar Errors & 5.60, 3.09 & 5.86, 3.54 \\
\bottomrule
\end{tabular}
\vspace{-5pt}
\caption{Activity, Entity F1 results of VHRED model vs. BPE-VHRED model tested on normal inputs.\label{tab:bpe-result}}
\vspace{-14pt}
\end{table}

\section{Related Works}
\label{sect:Related Works}
\noindent\textbf{Model-Dependent vs. Model-Agnostic Strategies:}
Many adversarial strategies have been applied to both Computer Vision~\cite{biggio2012poisoning,szegedy2013intriguing,goodfellow2014explaining,mei2015using,papernot2016limitations,narodytska2016simple,liu2016delving,carlini2017adversarial,papernot2017practical,mironenco2017examining,wong2017dancin,gao2018black} and NLP~\cite{jia2017adversarial,zhao2017generating,belinkov2017synthetic,shalyminov2017challenging,mironenco2017examining,iyyer2018adversarial}. Previous works have distinguished between 
model-aware strategies, where the adversarial algorithms have access to the model parameters, and 
model-agnostic strategies, where the adversary does not have such information~\cite{papernot2017practical,liu2016delving,narodytska2016simple}. We however, observed that within the 
model-agnostic
category, there are two subcategories. One is 
\textit{half-model-agnostic}, where although the adversary has no access to the model parameters, it is allowed to probe the target model and observe its output as a way to craft adversarial inputs~\cite{biggio2012poisoning,szegedy2013intriguing,goodfellow2014explaining,mei2015using,papernot2017practical,mironenco2017examining}. 
On the other hand, a 
\textit{pure-model-agnostic}
adversary, such as works by~\newcite{jia2017adversarial} and~\newcite{belinkov2017synthetic}, does not have any access to the model outputs when creating adversarial inputs, and is thus more generalizable across models/tasks. We adopt the pure-model-agnostic approach, only drawing inspiration from real-world noise, and testing them on the target model. 

\noindent\textbf{Adversarial in NLP:}
Text-based adversarial works have targeted both classification models~\cite{weston2015towards,jia2017adversarial,wong2017dancin,liang2017deep,samanta2017towards,shalyminov2017challenging,gao2018black,iyyer2018adversarial} and generative models~\cite{hosseini2017deceiving,henderson2017ethical,mironenco2017examining,zhao2017generating,belinkov2017synthetic}.
To our best knowledge, our work is the first to target generative goal-oriented dialogue systems with several new adversarial strategies in both Should-Not-Change and Should-Change categories, and then to fix the broken models through adversarial training (esp. using max-margin loss for Should-Change), and also achieving model robustness without using any adversarial data.

\section{Conclusion}
\label{sect:Conclusion}
We first revealed both the over-sensibility and over-stability of state-of-the-art models on Ubuntu and CoCoA dialogue tasks, via Should-Not-Change and Should-Change adversarial strategies. We then showed that training on adversarial inputs not only made the models more robust to the perturbations, but also helped them achieve new state-of-the-art performance on the original data (with further improvements when we combined strategies).
Lastly, we also proposed a BPE-enhanced VHRED model that not only trains faster with comparable performance, but is also robust to Grammar Errors even without adversarial training, motivating that if no strong adversary-generation tools (e.g., paraphraser) are available (esp. in low-resource domains/languages), we should try alternative model-robustness architectural changes.

\section*{Acknowledgments}
We thank the anonymous reviewers for their helpful comments and discussions. This work was supported by DARPA (YFA17-D17AP00022), Facebook ParlAI Research Award, Google Faculty Research Award, Bloomberg Data Science Research Grant, and Nvidia GPU awards. The views contained in
this article are those of the authors and not of the funding agency.

\appendix
\section{Appendix}

\subsection{Detailed Experimental Setup}
\label{sect:Detailed Experimental Setup}
\paragraph{Ubuntu Task Training Details}
We implemented VHRED and Reranking-RL in TensorFlow~\cite{abadi2016tensorflow} 
and employed greedy search for inference.
As shown in Table 1 (of the main paper), for both models we obtained Activity and Entity F1's higher than the VHRED results reported in~\newcite{serban2017multiresolution}.
Hence, each of these two implementations serves as a solid baseline for adversarial testing and training. 

\paragraph{Should-Not-Change Strategies}
For Random Swap, we allow up to $1$ swap of neighboring words per $4$ words in each utterance, 
and prevent already-swapped tokens from being transposed again. Each swap position is chosen randomly, except that we avoid swapping punctuations, end-of-utterance tokens (\textit{\_\_eou\_\_}), and end-of-turn tokens (\textit{\_\_eot\_\_}) since they mark the boundaries of sentences or clauses. 
For Stopword Dropout, we obtain a list of stopwords from the \textit{nltk.corpus} package~\cite{bird2004nltk},\footnote{We manually remove the pronouns, verbs, the \textit{5W1H} ("who", "what", "where", "when", "why" and "how") and "not" from the list because deleting them in an utterance usually incurs confusion.
}
and allow up to $8$ words to be dropped in each turn (which may contain several utterances).
For Data-level Paraphrasing, we only use the small version of PPDB 2.0,
since it contains paraphrase pairs with the highest precision/confidence.
For Generative-level Paraphrasing, we used the publicly available Pointer-Generator Networks
code\footnote{\url{https://github.com/becxer/pointer-generator}} and trained it on ParaNMT-5M.
Please refer to Table~\ref{tab:generative-para} for some random samples of the generated paraphrases.\footnote{We do not use word-overlap-based metrics to evaluate this model, because a perfect score would correspond to no change at all.}
For Grammar Errors, in addition to those extracted from the AESW dataset, we also add a heuristic where an inflected verb is replaced with its respective infinitive form,
and a plural noun with its singular form using the \textit{Pattern} package~\cite{smedt2012pattern}. These are common mistakes made by both native and non-native speakers~\cite{harap1930most,shaughnessy1979errors,chaney1999effect}. Note that we only keep an adversarial token if it is within the original vocabulary set, otherwise we would be converting the original token into an UNK token. The same applies to all other Should-Not-Change and Should-Change strategies.

\paragraph{Should-Change Strategies}
For Add Negation, we negate the first verb in each utterance.
For Antonym, we modify the first verb, adjective or adverb that has an antonym.

\paragraph{Human Evaluation}
To evaluate the effectiveness of adversarial training and testing, we conducted two human studies on MTurk. For each study, the utterances were randomly shuffled to anonymize model identity. Each assignment was annotated by one human evaluator located in the US, had an approval rate greater than $98\%$, and had at least $10,000$ approved HITs on record.

The first study compared the \textit{N-train + N-test} responses (corresponding to the score on the upper-right corner in Table 2 in the main paper) with each \textit{A-train + N-test} as well as the combined version of \textit{A-train + N-test} for the VHRED model, in order to evaluate how effective the adversarial trainings were. We adopted pairwise comparison (i.e., the annotators' task was to assign which one of the two responses wins on dialogue quality, or otherwise call it a tie) for $100$ samples for each adversarially trained model.

The second study juxtaposed original Ubuntu contexts with their paraphrases generated by the Pointer-Generator model (details in Section 3.1 of the main paper) to evaluate the quality of the Generative Paraphrasing strategy. To obtain a ceiling performance of the model, we also conducted a separate study where Turkers were presented with $100$ randomly sampled ground-truth ParaNMT-5M paraphrase pairs. For the Ubuntu experiment, we broke each context into utterances, and then randomly sampled $100$ such utterances and their paraphrases.\footnote{The average lengths of the utterances from Ubuntu and ParaNMT-5M are $13.11$ and $11.34$, respectively, indicating roughly fair comparison considering longer sentences are usually harder to paraphrase.} We adopted a five-point Likert scale, which corresponds to how good the paraphrase is.

\paragraph{CoCoA Task Training Details}
We directly adopted the publicly available code from~\newcite{he2017learning},\footnote{{\scriptsize\url{https://worksheets.codalab.org/worksheets/0xc757f29f5c794e5eb7bfa8ca9c945573/}}} and used their trained DynoNet model checkpoint to have reproduced all results.
For the Normal Inputs with Confusing Entity strategy, we randomly drew from the other bot's KB an entity of the same category (e.g., \textit{school}, \textit{company}, etc...) that the original one was in.

\subsection{Generative-Level Paraphrasing Examples}
\label{sect:Generative-Level Paraphrasing Examples}
We present some randomly sampled paraphrase examples in Table~\ref{tab:generative-para} (at the end of the paper). We can see that the Pointer-Generator Networks in general produce faithful paraphrases, though sometimes it skips some information from the source sequence.

\begin{table*}[h!]
\centering
\begin{tabular}{c|p{13.5cm}}
\toprule
Norm & just wondering how it runs . \\
Adv & he was just wondering how he works . \\
\midrule
Norm & you may need to create a new volume . \\
Adv & you should create a new volume . \\
\midrule
Norm & sorry i have no idea what that is \\
Adv & i do n't have any idea what that 's \\
\midrule
Norm & the installers partion manager tool is a bit . annoying . \\
Adv & the installers manager tool is a little annoying . \\
\midrule
Norm & had sound effects turned off in sound settings , didn't realize that controlled other applications . \\
Adv & the sound changed in the sound of the sound of the sound of the sound settings , didn't realized that controlled other applications . \\
\midrule
Norm & can i load an external hdd with ubuntu then use that to install it ? \\
Adv & can i have a external hdd with ubuntu then use it to install it ? \\
\midrule
Norm & are you using ubuntu on the pc right now , or working from windows ? \\
Adv & do you use ubuntu on the pc or working from windows ? \\
\midrule
Norm & download latest compat-wireless and compile . \\
Adv & download latest and compile . \\
\midrule
Norm & so now i'm going to try it with the snapshot version of the ppa instead and see if it crashes again . \\
Adv & i'm to try it with the snapshot of the ppa instead and see if it comes again . \\
\midrule
Norm & thanks , any idea on options to fix ? \\
Adv & you do n't know about options to fix ? \\
\bottomrule
\end{tabular}
\caption{Non-cherry picked examples for Generative-level Paraphrasing on the Ubuntu Dialogue Corpus. \textit{Norm} stands for Normal Inputs, while \textit{Adv} stands for adversarial inputs.}
\label{tab:generative-para}
\end{table*}

\begin{table*}[h!]
\centering
\begin{tabular}{p{9.3cm}|p{5.7cm}}
\toprule
\multicolumn{1}{c|}{Context} & \multicolumn{1}{|c}{Response} \\
\midrule
\textbf{N:} ... you could save your ubuntu files and reinstall Windows , then install ubuntu as a dual boot option \_\_eou\_\_ \_\_eot\_\_ aight buddy , so how do i get that **unknown** space back\_\_eou\_\_ \newline \textbf{Random Swap:} ... you could your save ubuntu and files Windows reinstall , then install ubuntu as dual a option boot \_\_eou\_\_ \_\_eot\_\_ aight buddy , so do how i that get space **unknown** back \_\_eou\_\_ & \textbf{NN:} you can use the Live CD , you can install Ubuntu on the same partition as the Windows partition \_\_eou\_\_ \newline \textbf{NA:} I am using ubuntu . \_\_eou\_\_ \newline \textbf{AA:} you can use Windows XP on the Windows partition , and then install Ubuntu on the same drive \_\_eou\_\_ \\

\midrule

\textbf{N:} ... ok then the problem is you don't have the nvidia driver \_\_eou\_\_ dpkg -l | grep nvidia \_\_eou\_\_ \_\_eot\_\_ i can see some nvidia thing :P \_\_eou\_\_ 
\newline
\textbf{Stopword Dropout:} ... ok then problem is you don't have nvidia driver \_\_eou\_\_ dpkg -l | grep nvidia \_\_eou\_\_ \_\_eot\_\_ i see nvidia thing :P
&
\textbf{NN:} sudo apt-get install nvidia-glx \_\_eou\_\_
\newline
\textbf{NA:} ok \_\_eou\_\_
\newline
\textbf{AA:} you can use the nvidia driver \_\_eou\_\_
\\

\midrule

\textbf{N:} ... if using hdmi , the display should be set properly . except perhaps for any overscan settings \_\_eou\_\_ \_\_eot\_\_ which package do i have to install for that ? \_\_eou\_\_
\newline
\textbf{Data-Level Para.:} ... if using hdmi , the displaying ought to be set correctly . except maybe for any overscan setting \_\_eou\_\_ \_\_eot\_\_ which packets do i have to installs for that ? \_\_eou\_\_
&
\textbf{NN:} you can use the nvidia-settings tool to install the drivers . \_\_eou\_\_
\newline
\textbf{NA:} the monitor is the monitor . \_\_eou\_\_
\newline
\textbf{AA:} you can use the nvidia-settings tool to see if the video card is installed . \_\_eou\_\_
\\

\midrule

\textbf{N:} at this point driver activated but not in use \_\_eou\_\_ \_\_eot\_\_ did you uninstall the old driver first ? \_\_eou\_\_
\newline
\textbf{Generative-Level Para.:} driver activated but not in use \_\_eou\_\_ \_\_eot\_\_ did you uninstall the old driver ?   \_\_eou\_\_
&
\textbf{NN:} i installed the driver from nvidia website \_\_eou\_\_
\newline
\textbf{NA:} ok \_\_eou\_\_
\newline
\textbf{AA:} i installed the driver from nvidia website \_\_eou\_\_
\\

\midrule

\textbf{N:} i would go with a thinkpad . they are pretty support and have decent hardware \_\_eou\_\_
\newline
\textbf{Grammar Errors:} ii would go with a thinkpad . they are quite support and have descent hardwares \_\_eou\_\_
&
\textbf{NN:} I know , but I don't want to buy a new laptop . \_\_eou\_\_
\newline
\textbf{NA:} Ubuntu . \_\_eou\_\_
\newline
\textbf{AA:} I know , but I don't want to buy a new laptop . \_\_eou\_\_
\\

\midrule

\textbf{N:} … suggest you reinstall grub . assuming no other changes , your root system and data should be present and safe . perhaps the bootloader scrambled somehow ? \_\_eou\_\_
\newline
\textbf{Add Negation:} … suggest you don't reinstall grub . assuming no other changes , your root system and data should be present and safe . perhaps the bootloader scrambled somehow ? \_\_eou\_\_
&
\textbf{NN:} I have a Windows XP partition , and I want to install Ubuntu on it . \_\_eou\_\_
\newline
\textbf{NA:} I have a Windows XP partition , and I want to install Ubuntu on it . \_\_eou\_\_
\newline
\textbf{AA:} I don't know how to do that , but I don't know how to do that . \_\_eou\_\_
\\

\midrule

\textbf{N:} 100\% of your RAM . You have it to be used . \_\_eou\_\_ \_\_eot\_\_ what what ? \_\_eou\_\_
\newline
\textbf{Antonym:} 100\% of your RAM . You refuse it to be used . \_\_eou\_\_ \_\_eot\_\_ what what ? \_\_eou\_\_
&
\textbf{NN:} you have a lot of RAM . \_\_eou\_\_
\newline
\textbf{NA:} you have a lot of RAM . \_\_eou\_\_
\newline
\textbf{AA:} I don't know . I don't use it much . \_\_eou\_\_
\\

\bottomrule
\end{tabular}
\caption{Selected examples before and after adversarial training for the VHRED model with each strategy.}
\label{tab:selected-examples-VHRED}
\end{table*}

\begin{table*}[h!]
\centering
\begin{tabular}{p{9.3cm}|p{5.7cm}}
\toprule
\multicolumn{1}{c|}{Context} & \multicolumn{1}{|c}{Response} \\
\midrule

\textbf{N:} I didn't install LUKS and honestly don't know what it is ;) \_\_eou\_\_ \_\_eot\_\_ Good ... one less thing to consider ! ... So the **unknown** are NTFS ... how are they being shared over the network if not NFS ? Samba ? \_\_eou\_\_
\newline
\textbf{Random Swap:} I install didn't LUKS and honestly don't what know it ;) is \_\_eou\_\_ \_\_eot\_\_ Good ... one thing less consider to ! ... So the are **unknown** ... NTFS are how being they shared the over network not if NFS ? Samba ? \_\_eou\_\_
&
\textbf{NN:} I don't know , I don't use samba . \_\_eou\_\_
\newline
\textbf{NA:} I don't know , I don't know . I don't know what you mean by " mount " \_\_eou\_\_
\newline
\textbf{AA:} I don't know , I just installed samba on my own machine and it worked fine . \_\_eou\_\_
\\

\midrule

\textbf{N:} ... try to get a broadcom 43xx chip and it'll be a simple install \_\_eou\_\_
\newline
\textbf{Stopword Dropout:} ... try get broadcom 43xx chip it'll be simple install \_\_eou\_\_
&
\textbf{NN:} i have a broadcom 4311 \_\_eou\_\_
\newline
\textbf{NA:} ok \_\_eou\_\_
\newline
\textbf{AA:} i have a broadcom card \_\_eou\_\_
\\

\midrule

\textbf{N:} wine should install winetricks automatically \_\_eou\_\_ \_\_eot\_\_ okay \_\_eou\_\_
\newline
\textbf{Data-Level Para.:} wine ought installs winetricks automatically \_\_eou\_\_ \_\_eot\_\_ ok \_\_eou\_\_
&
\textbf{NN:} you can install wine from the repos \_\_eou\_\_
\newline
\textbf{NA:} it 's a bit of a pain in the ass \_\_eou\_\_
\newline
\textbf{AA:} you can install wine from the repos \_\_eou\_\_
\\

\midrule

\textbf{N:} You can find xorg . conf in /etc/X11 . It 's not needed unless it is . ;-) You may need to create one yourself . \_\_eou\_\_
\newline
\textbf{Generative-Level Para.:} You may find xorg . conf in /etc/X11 . It 's not necessary until it is . You may be required to create one . \_\_eou\_\_
&
\textbf{NN:} i have a backup of my xorg . conf \_\_eou\_\_
\newline
\textbf{NA:} i know , but i don't know what to do with it \_\_eou\_\_
\newline
\textbf{AA:} i have a backup of my xorg . conf file , but i can't find it in /etc/X11 \_\_eou\_\_
\\

\midrule

\textbf{N:} ... but i know name of package , how to install it then \_\_eou\_\_
\newline
\textbf{Grammar Errors:} ... but ii know names of packaging , how to installs it then \_\_eou\_\_
&
\textbf{NN:} you can install it from the repos \_\_eou\_\_
\newline
\textbf{NA:} no idea \_\_eou\_\_
\newline
\textbf{AA:} you can install it from synaptic \_\_eou\_\_
\\

\midrule

\textbf{N:} … **unknown** is there .... but they wont execute \_\_eou\_\_ bash is saying no such file or dir \_\_eou\_\_ \_\_eot\_\_ ls -l **unknown** \_\_eou\_\_
\newline
\textbf{Add Negation:} … **unknown** isn't there .... but they wont execute \_\_eou\_\_ bash isn't saying no such file or dir \_\_eou\_\_ \_\_eot\_\_ ls -l **unknown** \_\_eou\_\_
&
\textbf{NN:} -rwxr-xr-x 1 root root root 4096 Nov 2008 \_\_eou\_\_
\newline
\textbf{NA:} -rwxr-xr-x 1 root root root 4096 Nov 2008 \_\_eou\_\_
\newline
\textbf{AA:} bash : bash : bash : command not found \_\_eou\_\_
\\

\midrule

\textbf{N:} depends on the fs of the sd card and what device is it .. \_\_eou\_\_ \_\_eot\_\_ the filesystem is FAT32 \_\_eou\_\_
\newline
\textbf{Antonym:} depends off the fs of the sd card and what device is it .. \_\_eou\_\_ \_\_eot\_\_ the filesystem is FAT32 \_\_eou\_\_
&
\textbf{NN:} then you can use the usb drive to mount it \_\_eou\_\_
\newline
\textbf{NA:} then you can use the usb drive to mount it \_\_eou\_\_
\newline
\textbf{AA:} then it 's not mounted .. \_\_eou\_\_
\\

\bottomrule
\end{tabular}
\caption{Selected examples before and after adversarial training for the Reranking-RL model with each strategy.}
\label{tab:selected-examples-RL}
\end{table*}

\subsection{Adversarial Training Details}
Observing that adversarial trainings on the Should-Change strategies in some cases improve the model's performance on the original data (Table 3 of the main paper), we employ a max-margin baseline as a sanity check to see whether the enhancement comes from the max-margin objective itself or is due to the adversarial data. For that, we feed random source inputs from the Ubuntu dataset in place of the adversarial inputs, and use exactly the same approach to train the model. We found that there was no stat. significant change in the model's performance for both VHRED and Reranking-RL, either starting from a converged model or from scratch.

\subsection{Qualitative Examples Before and After Adversarial Training}
\label{sect:Qualitative Examples before and after Adversarial Training}
We next present selected examples before and after adversarial training for both VHRED and Reranking-RL with each strategy (Table~\ref{tab:selected-examples-VHRED} and~\ref{tab:selected-examples-RL}, at the end of the paper), including the one already shown in the main paper (see Table 7 of the main paper).

\subsection{Sampling-based Max-Margin Loss}
While the ground-truth-based max-margin approach does work if we train with max-margin loss from scratch (verified through our preliminary experiments), it will not work if the model has already been well-pretrained, because it would be trivially true that the margin $\log P(t_i|s_i) - \log P(t_i|a_i) > 0$. This is an issue because it takes too much time to train the model from scratch for each strategy.
To address this challenge, we propose a novel max-margin loss that can work even with pretrained models. The idea is to replace both $t_i$ (target sequence) in $L_{\textsc{mm}}$ (Eq. 3 in the main paper) with $g_i$ (greedy-decoded sequence).
By treating the greedy-decoded sequences as targets, we are essentially introducing fresh training examples so that $\log P(g_i|s_i) - \log P(g_i|a_i) > 0$ will not be trivially true. In addition, the greedy-decoded sequences keep evolving over time, bringing in more previously unseen negative examples for the model to train on.

Note that we do not perform adversarial training with max-margin loss for the three Should-Change strategies on the CoCoA task, because for Random Inputs and Random Inputs with Kept Entities, both $P(t_i|a_i)$ and $P(g_i|a_i)$ will be close to zero since the source sequences consist of random tokens. This makes the margin trivially large, resulting in almost no model parameter updates during training. For Normal Inputs with Confusing Entities, adversarial training would encourage a bot to ignore entities mentioned by the other bot when these entities are in its own KB, because the bot is agnostic to whether the mentioned entities are in the actually shared entry or the confusing entries, even during training. This is the opposite of what we intend the model to behave.

\bibliography{references}
\bibliographystyle{acl_natbib_nourl}

\end{document}